\documentclass{article}

\usepackage{arxiv}

\usepackage[utf8]{inputenc} 
\usepackage[T1]{fontenc}    
\usepackage{hyperref}       
\usepackage{amsmath,amssymb,amsfonts}
\usepackage{url}            
\usepackage{booktabs}       
\usepackage{amsfonts}       
\usepackage{nicefrac}       
\usepackage{microtype}      
\usepackage{cleveref}       
\usepackage{lipsum}         
\usepackage{graphicx}
\usepackage{doi}
\usepackage{comment}
\usepackage[ruled,vlined]{algorithm2e}
\usepackage[backend=bibtex, isbn=false, url=false, sorting=none]{biblatex}
\title{AutoGCN - Towards Generic Human Activity Recognition with Neural Architecture Search}

\date{}

\newif\ifuniqueAffiliation
\uniqueAffiliationtrue

\ifuniqueAffiliation 
\author{ \href{https://orcid.org/0009-0005-6310-408X}{\includegraphics[scale=0.06]{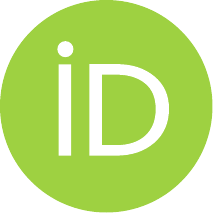}\hspace{1mm}Felix Tempel}\\
	Faculty of Informatics\\
	Norwegian University of Science and Technology\\
	Trondheim, Norway \\
	\texttt{felix.e.f.tempel@ntnu.no} \\
	\And
	\href{https://orcid.org/0000-0003-1820-6544}{\includegraphics[scale=0.06]{orcid.pdf}\hspace{1mm}Inga Strümke} \\
	Faculty of Informatics\\
	Norwegian University of Science and Technology\\
	Trondheim, Norway \\
	\texttt{inga.strumke@ntnu.no} \\
	\AND
    \href{https://orcid.org/0000-0002-2469-1809}{\includegraphics[scale=0.06]{orcid.pdf}\hspace{1mm}Espen Alexander F. Ihlen} \\
	Faculty of Medicine and Health Sciences\\
	Norwegian University of Science and Technology\\
	Trondheim, Norway \\
	\texttt{espen.ihlen@ntnu.no} \\
}
\else
\usepackage{authblk}

\newbox{\orcid}\sbox{\orcid}{\includegraphics[scale=0.06]{orcid.pdf}} 
\author[1]{%
	\href{https://orcid.org/0000-0000-0000-0000}{\usebox{\orcid}\hspace{1mm}David S.~Hippocampus\thanks{\texttt{hippo@cs.cranberry-lemon.edu}}}%
}
\author[1,2]{%
	\href{https://orcid.org/0000-0000-0000-0000}{\usebox{\orcid}\hspace{1mm}Elias D.~Striatum\thanks{\texttt{stariate@ee.mount-sheikh.edu}}}%
}
\affil[1]{Department of Computer Science, Cranberry-Lemon University, Pittsburgh, PA 15213}
\affil[2]{Department of Electrical Engineering, Mount-Sheikh University, Santa Narimana, Levand}
\fi

\hypersetup{
pdftitle={AutoGCN},
pdfsubject={q-bio.NC, q-bio.QM},
pdfauthor={Felix Tempel},
pdfkeywords={First keyword, Second keyword, More},
}

\begin{document}

\maketitle

\begin{abstract}
This paper introduces AutoGCN, a generic Neural Architecture Search (NAS) algorithm for Human Activity Recognition (HAR) using Graph Convolution Networks (GCNs).
HAR has enjoyed increased attention due to advances in deep learning, increased data availability, and enhanced computational capabilities. 
Concurrently, GCNs have shown promising abilities in modeling relationships between body key points in a skeletal graph. 
Typically, domain experts develop dataset-specific GCN-based methods, which limits their applicability beyond the specific context. 
AutoGCN seeks to address this limitation by simultaneously searching for the ideal hyperparameters and architecture combination within a versatile search space using a reinforcement controller while balancing optimal exploration and exploitation behavior with a knowledge reservoir during the search process.
We conduct extensive experiments on two large datasets focused on skeleton-based action recognition to assess the proposed algorithm's performance.
Our experimental results demonstrate the effectiveness of AutoGCN in constructing optimal GCN architectures for HAR, outperforming conventional NAS and GCN methods, as well as random search.
These findings highlight the significance of a diverse search space and an expressive input representation to achieve good model performance and generalizability.
\end{abstract}


\section{Introduction}
\label{sec:introduction}
Human Activity Recognition (HAR), the process of identifying and categorizing human activities, has witnessed a surge in interest in recent years. 
This momentum is attributed to advancements in deep learning techniques, the increased availability of data, and enhanced computational capabilities \cite{fengSkeletonGraphNeuralNetworkBasedHuman2022}. 
HAR has applications in a wide range of domains, including video surveillance, human-computer interaction, and healthcare, highlighting its significance and broad impact \cite{fengSkeletonGraphNeuralNetworkBasedHuman2022, groosDevelopmentValidationDeep2022, vrigkasReviewHumanActivity2015, arshadHumanActivityRecognition2022}.
Within the realm of HAR, the models aim to capture representations of humans using a skeleton graph built of body key points \cite{fengSkeletonGraphNeuralNetworkBasedHuman2022}. 
In the context of HAR, the traditional feature extraction method relies on manually crafted descriptors to capture fundamental attributes \cite{yangEigenJointsbasedActionRecognition2012, bobickRecognitionHumanMovement2001}. 
Another approach involves formulating the problem as a deep learning task in Euclidean space. 
This approach uses the capabilities of Convolutional Neural Networks and Recurrent Neural Networks associated with recognition tasks.
However, approaches involving these architectures have a significant shortcoming in modeling the inherent structures in skeletons, as the input data is serialized \cite{fengSkeletonGraphNeuralNetworkBasedHuman2022}.
Recently, Graph Neural Networks (GNNs), specifically Graph Convolution Networks (GCNs), have emerged as a cutting-edge solution in HAR \cite{yanSpatialTemporalGraph2018}. 
GCNs have achieved remarkable performance by leveraging the relationships between different segments and joints, treating these as nodes and edges in an adjacency matrix, making convolutional operations possible \cite{fengSkeletonGraphNeuralNetworkBasedHuman2022, yanSpatialTemporalGraph2018}.

While GCN-based methods have shown promising results, they are primarily developed by domain experts and often tailored to specific datasets through a trial-and-error process. 
This limits their potential impact, as applying them to a broader range of problems can be challenging or when domain knowledge is unavailable.
One approach to overcoming those challenges is automatically building task-specific deep learning models using Neural Architecture Search (NAS) \cite{zophNeuralArchitectureSearch2017}.
By employing NAS, researchers can discover optimal architectures from a defined search space through a search procedure, enhancing the overall performance and generalizability of the resulting models.

In this work, we propose a generic NAS algorithm, AutoGCN, that dynamically learns both hyper- and architecture parameters to construct a framework that maximally fits the HAR task.
We present the following contributions:

\begin{itemize}
\item A generic approach to construct a GCN architecture for HAR with NAS, which can concurrently optimize hyperparameters and the model architecture. Additionally, the approach can be applied to different datasets within HAR and performs better than random search and other NAS procedures for GCNs.
\item An update mechanism incorporating a knowledge reservoir to balance exploration and exploitation during the search process, leading to an accelerated and efficient search procedure while ensuring optimal exploration behavior.
\item A diverse search space construction that enables the usage of different input representations to enhance the network performance and generalizability further.
\end{itemize}

\section{Related Work}
\subsection{Skeleton-based HAR}
\noindent HAR within the domain of skeletal data attracts increasing interest due to its rectified robustness against background noise, occlusion, or scaling implications when compared to conventional RGB methods \cite{vrigkasReviewHumanActivity2015}.
Since the nature of the skeleton data modality is non-euclidean, current approaches commonly use GCNs, as these can operate on unstructured data.
The first work in this field stems from \cite{yanSpatialTemporalGraph2018}, introducing ST-GCN to learn spatial and temporal patterns from the input data with a tailored GCN layer. 
This foundation makes the handcrafted feature engineering obsolete while achieving superior performances compared to previous methods \cite{yanSpatialTemporalGraph2018}.
Several subsequent studies have adopted this technique and introduced multi-stream architectures to incorporate different kinematic variables from which the model can learn.
A two-stream adaptive graph convolutional network named 2s-AGCN was introduced in \cite{shiTwoStreamAdaptiveGraph2019} to consider the skeleton's second-order information and learn the graph's topology.
In a later study \cite{Song2022ConstructingSA}, Efficient-GCN was introduced, built upon three input branches and incorporating techniques from separable convolution to make the network more efficient.
While these approaches yield good performances, they are all designed by human experts, limiting their generalizability to scenarios beyond the tested databases or usage by non-domain experts. 
This leaves the problem of the automatic construction of generic models unresolved.

\subsection{Neural Architecture Search}
\noindent For most tasks, deep neural networks must have a customized architecture to accomplish their objectives effectively. 
Initially, deep neural networks were used to automate the manual feature engineering aspect of machine learning. 
The field has since progressed further with the introduction of NAS and automated machine learning (AutoML), which automates the construction of neural architectures and accompanying tasks like hyperparameter optimization, e.g., batch size, learning rate, or algorithm selection \cite{hutterAutomatedMachineLearning2019}.
These approaches have already given rise to architectures in the area of image classification that outperform models designed by human experts, as highlighted in studies such as \cite{zophNeuralArchitectureSearch2017, whiteNeuralArchitectureSearch2023}.

Zoph et al. made significant contributions to the field of NAS with their pioneering work \cite{zophNeuralArchitectureSearch2017}. 
Their approach introduced a reinforcement learning algorithm searching for optimal architectures, yielding competitive performance on the CIFAR-10 and Penn Treebank benchmarks \cite{krizhevsky2009learning, marcus-etal-1993-building}. 
However, it is worth noting that this method employed a substantial amount of computational resources, utilizing 800 GPUs over two weeks  \cite{whiteNeuralArchitectureSearch2023}. 
Significant research has focused on improving the efficiency of NAS approaches to address the first approach's resource-intensive aspect.
For example, in the work that developed the Differentiable Architecture Search (DARTS) technique \cite{liuDARTSDifferentiableArchitecture2019}, a gradient-based optimization methodology was presented, enabling efficient exploration of designs by utilizing a continuous relaxation of the search space. 
Another influential study, \cite{phamEfficientNeuralArchitecture2018}, introduced the Efficient Neural Architecture Search (ENAS) approach, which addressed the computational inefficiency of NAS through parameter sharing across child models, yielding greater efficiency and scalability. 
The Progressive Neural Architecture Search (PNAS) \cite{liuProgressiveNeuralArchitecture2018} concept established a step-by-step expansion strategy to shorten the search process while maintaining competitive performance. 
Furthermore, the NASNet technique \cite{zophLearningTransferableArchitectures2018} focused on identifying transferable architectural building blocks that perform well across various tasks and datasets.

A new approach named AutoHAS incorporating hyperparameters and architecture components in the search was introduced in \cite{dongAutoHASEfficientHyperparameter2021}. 
This comprehensive method aims to optimize the architectural design and the associated hyperparameters like the chosen optimizer and its learning rate for increased performance.
Therefore, the shared network weights and a reinforcement learning controller are updated alternately to find the best probability distribution in the combined search space, resulting in the best-performing architecture and hyperparameter configuration.
By extending the framework to incorporate hyperparameter optimization, AutoHAS went beyond the traditional scope of NAS and encompassed a broader spectrum of optimization possibilities.
The inclusion of hyperparameter search is a significant enhancement as it allows for the tuning of various factors that impact model performance, such as learning rates, regularization parameters, and batch sizes, which have a significant impact on the model performance \cite{whiteNeuralArchitectureSearch2023}.
In contrast, AutoHAS relies on the straightforward Reinforce algorithm \cite{suttonPolicyGradientMethods1999}, which overlooks previously sampled and trained candidate architectures during the update procedure, resulting in an imbalance between exploration and exploitation behaviors. Additionally, the architectural search is confined to basic building blocks, and the exploration of hyperparameters is restricted to selecting the most suitable learning rate and optimizer.

To date, NAS research has primarily focused on and applied to image classification tasks. 
In the domain of HAR GCNs, the NAS research focus has been limited to \cite{pengLearningGraphConvolutional2020, jiangSkeletonBasedHumanAction2023}, to the best of our knowledge. 
In the work of \cite{pengLearningGraphConvolutional2020}, NAS was employed to discover the most effective architecture by approximating spatiotemporal cues and leveraging Chebyshev polynomials of varying orders. 
This approach involved constructing a search space consisting of three dynamic graph substructures: spatial, temporal, and spatiotemporal. 
Additionally, the authors searched for eight function modules that were subsequently applied to each network layer.
Expanding upon these concepts, SNAS-GCN represents an extension of this research, optimizing the search space and implementing a single-path one-shot approach for improved search efficiency \cite{jiangSkeletonBasedHumanAction2023}. 
The experimental findings of SNAS-GCN demonstrate a reduction in search time compared to the previous work \cite{pengLearningGraphConvolutional2020}, albeit with a diminished level of accuracy.
Despite the significant contributions of \cite{pengLearningGraphConvolutional2020, jiangSkeletonBasedHumanAction2023} to the field of skeleton GCNs through utilizing NAS, their work has certain limitations.
One limitation lies in their reliance on approximating spatiotemporal cues and utilizing Chebyshev polynomials, which can lead to information loss and impact the derived architectures' accuracy and robustness.
Furthermore, the search for a modest set of eight function modules applied to the network layers may restrict the expressiveness and flexibility of the discovered architectures.
The lack of flexibility in the network architecture is a further notable limitation, as it is fixed to ten layers with uniform channel sizes and composition.
This rigidity restricts the expressiveness of the architecture, potentially hindering its ability to capture complex patterns in the data effectively.
In addition, the authors only search for the spatial and temporal filters solely on the joint data while sharing them with the other four dataset modalities within NTU RBG+D \cite{shahroudyNTURGBLarge2016}, and, as a result of this, overlook the potential variance present in these modalities.
Moreover, the final performance of the architecture relies on an ensemble score derived from two distinct data modalities, making it necessary to use more resources to search and train those models.
Considering these limitations, a more extensive and diverse search space combined with an ample data representation, the model could leverage a richer set of operations and adapt better to the specific task requirements, potentially leading to improved performance.

\section{AutoGCN}
\begin{figure}[hbtp]
    \centering
    \includegraphics[width=0.5\textwidth]{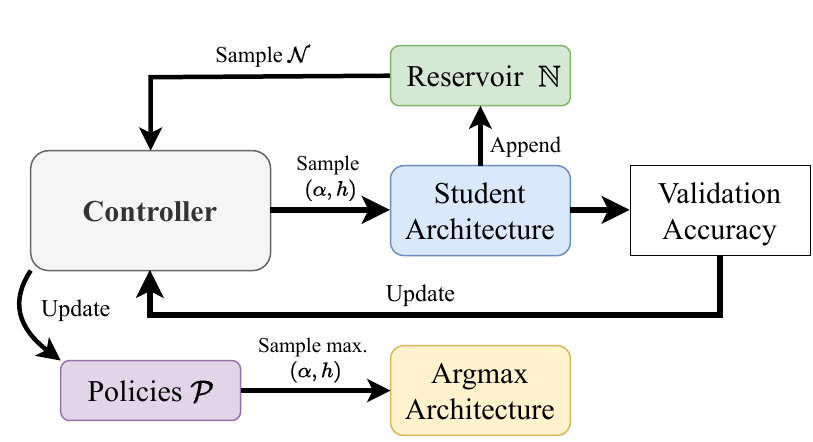}
    \caption{Overview of the proposed AutoGCN algorithm.}
\label{fig:autogcn}
\end{figure}

\noindent AutoGCN is a novel NAS framework that predicts action classes for a given sequence of skeletons. 
In this section, we provide AutoGCN's technical details. 
First, we introduce some preliminaries and the data representation as the foundational input features. 
Second, we discuss the defined search space and its implications. 
Third and finally, we explain the search algorithm in detail.
Fig. \eqref{fig:autogcn} illustrates the workflow in the AutoGCN framework.

\subsection{Preliminaries}
\noindent The human skeleton is modeled as a spatiotemporal graph by connecting the spatial graph of each time frame along a defined temporal dimension $d$, enabling the network to process spatial and temporal information effectively. Therefore, the skeleton sequence is transformed into an undirected graph described as $G = (V, E, \mathbf{A})$, where $n = |V|$ denotes the nodes connected by $|E|$ edges. These attributes are represented in an adjacency matrix $\mathbf{A} \in \mathcal{R}^{n \times n}$. 

\subsection{Data representation}
\noindent Prior research on skeleton-based action recognition demonstrated the paramount importance of data preprocessing \cite{chenChannelwiseTopologyRefinement2021, liSkeletonBasedActionRecognition2019, nieViewInvariantHumanAction2019, qinMultiStagePart2022, shiTwoStreamAdaptiveGraph2019, zhangMultiScaleSemanticsGuidedNeural2021, zhangSemanticsGuidedNeuralNetworks2020}. 
In the context of this study, the input features are categorized into four distinct groups, namely: \textbf{1)} position $P$, \textbf{2)} velocity $V$, \textbf{3)} bone $ B \in \{L, \beta\}$, and \textbf{4)} acceleration features $A$, corresponding to the fundamental kinematic properties of the human body \cite{zatsiorskyKinematicsHumanMotion1998}. 
The input sequence $\mathcal{X}$, denoted as $\mathcal{X} \in \mathcal{R}^{C_{in} \times T_{in} \times V_{in} \times M_{in}}$ represents a sequence of skeleton data inputs, where $C_{\text{in}}$ denotes the number of channels indicating the $\{x, y, z\}$ orientation, $T_{\text{in}}$ represents the number of frames in the sequence, $V_{\text{in}}$ specifies the number of vertices representing spatial points, and $M_{\text{in}}$ depicts the number of skeletons present in the sequence.
This input sequence undergoes a subsequent partitioning into the mentioned categories.
The relative position $P$ is calculated via
\begin{equation}
	\label{eq:position}
	\forall i = 1, 2, \ldots, V_{in}, \quad P = x[:, :, i, :] - x[:, :, c, :],
\end{equation}
where $c$ is the chosen central joint.
The velocity \(V\) is determined via
\begin{equation}
	\label{eq:vel}
	\forall t = 1, 2, \ldots, T_{in}, \quad V = \frac{{x[:, :, t + d, :] - x[:, :, t, :]}}{d}.
\end{equation}
The parameter \(d\) determines the timeframes over which the velocity is measured, with different values yielding different temporal spans. 
In this study, the range of \(d\) is within \([1,2]\).
To calculate the acceleration $A$, the velocity from \eqref{eq:vel} is differentiated with respect to time. 
Since the equation involves discrete time steps, finite differences are used to approximate the derivative.
The acceleration $A$ is then calculated as follows
\begin{alignat}{2}
    \label{eq:acc}
		& \forall t = 1, 2, \ldots, T_{in}, && A =	\frac{d V}{d t}, \\
		& \forall t = 1, 2, \ldots, T_{in} - d, \quad  && A = \frac{{V[:, :, t+d,:] - V[:, :, t,:]}}{{d}}.
\end{alignat}
An 8th-order Butterworth filter is also introduced to compensate by the derivative \cite{crennaFilteringBiomechanicalSignals2021} from the acceleration feature.
This meticulous filtering process is critical in enhancing the signal quality, enabling a more accurate and reliable representation of the underlying data.
The bone features $B$ are divided into bone length $L$ and respective angles $\beta$ which have been introduced in \cite{Song2022ConstructingSA}:
\begin{alignat}{2}
		\label{eq:bone}
		& \forall i = 1, 2, \ldots, V_{in}, \quad && L = x[:, :, i, :] - x[:, :, i_{adj}, :], \\
		& \forall j = 1, 2, \ldots, V_{in}, && \beta = cos^{-1} (\frac{l_{j, w}}{\sqrt{l^{2}_{j,x} + l^{2}_{j,y} + l^{2}_{j,z} }}).
\end{alignat}
Here $i_{adj}$ is the adjacent joint, while $w \in \{x,y,z\}$ stands for the three orientation coordinates of the skeleton.

\subsection{Spatial Graph Convolution}
\noindent Yan et al. \cite{yanSpatialTemporalGraph2018} pioneered the application of graph convolution in the domain of skeleton action recognition. 
The following relation encapsulates the essence of their approach:

\begin{equation}
\label{eq:graphconv}
f_{out}(v_{ti}) = \sum_{v_{tj} \in B(v_{ti})} \frac{1}{Z_{ti}(v_{tj})} f_{in} (v_{tj}) \mathbf{w} (l_{ti} (v_{tj})).
\end{equation}
Here, $v_{ti}$ represents the $i$-th joint at the $t$-th frame, and $f_{out}(.)$ and $f_{in}(.)$ correspond to the output and input features of the respective joints. 
Furthermore, $B(v_{ti})$ denotes the set of neighboring joints of $v_{ti}$, while $B_{v_{ti}}$ defines the convolutional sampling area, specifically encompassing vertexes that are $1$-distance neighbors (i.e., immediate neighboring joints). 
The normalization factor $Z_{ti}(v_{tj})$ represents the size of the corresponding subset, and its purpose is to equalize the influence of various subsets on the final output.
The weighting function $\mathbf{w(.)}$ incorporates the mapping function $l_{ti}(.)$, which serves to establish diverse partitioning strategies, such as uni-labeling, distance, and spatial configuration partitioning \cite{yanSpatialTemporalGraph2018}.
To further refine the formulation, Equation \eqref{eq:graphconv} is written as
\begin{equation}
	\label{eq:graphconv_2}
	\mathbf{f}_{out} =  \sum_{j} \mathbf{W}_{j} \mathbf{f}_{in} (\mathbf{\Lambda}_{j}^{-\frac{1}{2}} \mathbf{A}_{j} \mathbf{\Lambda}_{j}^{-\frac{1}{2}} \mathbf{M}_{j}) \,.
\end{equation}
Here, $\mathbf{f}_{out}$ denotes the output features, while $\mathbf{W}_{j}$, $\mathbf{\Lambda}_{j}$, $\mathbf{A}_{j}$, and $\mathbf{M}_{j}$ contribute to the weighting, diagonal degree, adjacency, and normalization matrices, respectively. This transformation enables an enhanced representation of the skeleton sequences using graph convolution techniques \cite{yanSpatialTemporalGraph2018}.

\subsection{Search Space}
\noindent Diverging from conventional image-based NAS algorithms, which typically incorporate a limited set of elementary building blocks, such as convolutional layers with varying kernel sizes, skip connections, or max pooling layers \cite{dongNASBench201ExtendingScope2020, yingNASBench101ReproducibleNeural2019}, AutoGCN adopts a distinctive approach. 
Recognizing that these simplistic blocks cannot adequately capture the data representation of a skeletal sequence, AutoGCN embraces the integration of prior knowledge. 
This algorithm aims to exploit the full potential of skeletal sequence information by defining the search space using architectural components known to be suitable for skeletal data \cite{chenChannelwiseTopologyRefinement2021, huSpatialTemporalGraph2022, pengLearningGraphConvolutional2020, phamEfficientNeuralArchitecture2018, shiTwoStreamAdaptiveGraph2019, yanSpatialTemporalGraph2018}. 
This departure from traditional methodologies acknowledges the need for a more nuanced and informed exploration of possible architectures in this domain \cite{whiteNeuralArchitectureSearch2023, liuDARTSDifferentiableArchitecture2019}.
In addition to incorporating prior knowledge in the search space, AutoGCN also encompasses hyperparameters like optimizer type, momentum, batch size, L2 regularisation, and learning rate. 
Building upon \cite{dongAutoHASEfficientHyperparameter2021}, the hyperparameter search space can be formulated as
\begin{equation}
	h =  \sum_{i=1}^{m} C_{i}^{h} \mathcal{B}_{i} \; s.t. \;  \sum_{i=1}^{m} C_{i}^{h} =1, \; 
    \mathrm{and} \; C_{i}^{h} \in 
    \begin{cases}
        [0, 1],       &  \text{if cont.} \\
        \{0,1\},  &  \text{if categ.},
    \end{cases}
\end{equation}

where the hyperparameter \(h\) is a sum of weighted components \(C_i^h\) of the predefined values \(\mathcal{B}_i\), satisfying two conditions: the weights sum up to 1, to ensure these are normalized and that the combined influence of all components adds up to the final hyperparameter value, and \(C_i^h\) behaves either as a continuous value in \([0, 1]\) or as a binary choice  \{0,1\} depending on whether it represents a continuous or categorical hyperparameter.

The architectural search space can be encapsulated in the following equation \cite{dongAutoHASEfficientHyperparameter2021}
\begin{equation}
    \alpha = \sum_{i=1}^{n} \sum_{j=1}^{k} C_{i, j}^{\alpha} \Delta_{i,j}  \; s.t. \; \sum_{i=1}^{n} \sum_{j=1}^{k} C_{i,j}^{\alpha}=1, 
        \mathrm{and} \; C_{i,j}^{\alpha} \in 
        \begin{cases}
            [0, 1],       &  \text{if cont.} \\
            \{0,1\},  &  \text{if categ.},
        \end{cases}
\end{equation}

where $n$ defines the chosen architecture type $C_{i,j}^{\alpha}$ and $k$ the corresponding value out of the predefined set of architectural parameters $\Delta_{i,j}$. 
Just as with the hyperparameters, the architecture can be categorical or continuous.
By considering both the architectural components and the hyperparameters, AutoGCN ensures a comprehensive search process, allowing for the fine-tuning of multiple parts to achieve an optimal outcome.
Table \eqref{tab:search_space} shows the entire search space and its variables.

\begin{table}[]
\centering
\caption{Search Space compromising the parameter and the corresponding value ranges, grouped into the effective areas.}
\label{tab:search_space}

\begin{tabular}{@{}ll@{}}
\toprule
\textbf{Parameter}        & \textbf{Range}                                                     \\ \midrule
\multicolumn{2}{l}{\textbf{Common}}                                             \\ \midrule
Activation layer        & {[}Relu, Relu6, Hardswish, Swish{]}               \\
Attention layer         & {[}Stja, Ca, Fa, Ja, Pa{]} \cite{Song2022ConstructingSA}                \\
Conv. layer type        & {[}Basic, Bottleneck\cite{Song2022ConstructingSA}, Sep\cite{Song2022ConstructingSA} \\
                        & SG\cite{Song2022ConstructingSA}, V3\cite{howardSearchingMobileNetV32019}, Shuffle\cite{zhangShuffleNetExtremelyEfficient2017}{]} \\
Dropout probability & {[}0.15, 0.2, 0.25, 0.3{]}                                \\ 
Init layer size      & {[}64, 128, 156, 195, 256{]}                                    \\ \midrule
\multicolumn{2}{l}{\textbf{Input stream}}                                             \\ \midrule
Blocks in        & {[}1, 2, 3{]}                                                  \\
Depth in         & {[}1, 2, 3{]}                                                \\
Stride in        & {[}1, 2, 3{]}                                             \\
Scaling factor	 & {[}0.4, 0.6, 0.8{]}										\\
Temporal window  & {[}3, 5, 7{]}                                             \\
Graph distance   & {[}2, 3, 4{]}                                             \\
Reduction factor & {[}1.225, 1.25, 1.275, 1.3, 1.325, 1.35{]}                \\ \midrule
\multicolumn{2}{l}{\textbf{Main stream}}                                              \\ \midrule
Blocks main      & {[}3, 4, 5, 6{]}                                       \\
Depth main       & {[}1, 2, 3, 4{]}               							\\
Graph distance   & {[}7, 9, 11{]}                                            \\
Shrinkage        & {[}1, 2, 4, 6{]}                                          \\
Residual layer   & {[}True, False{]}                                         \\
Adaptive         & {[}True, False{]}                                         \\ \midrule
\multicolumn{2}{l}{\textbf{Optimizer}}                                                \\ \midrule
Optimizer        & {[}SGD, Adam, AdamW{]}                                       \\ 
Learning rate    & {[}0.1, 0.05, 0.01{]}                                     \\
Weight decay     & {[}0.0, 0.01, 0.001, 0.0001{]}							\\
Momentum     	 & {[}0.5, 0.9, 0.99{]}									\\
Batch size    	 & {[}8, 16, 24{]}											\\ \bottomrule

\end{tabular}

\end{table}

\subsection{Architecture}
\begin{figure*}[hbt]
	\centering
	\includegraphics[width=\textwidth]{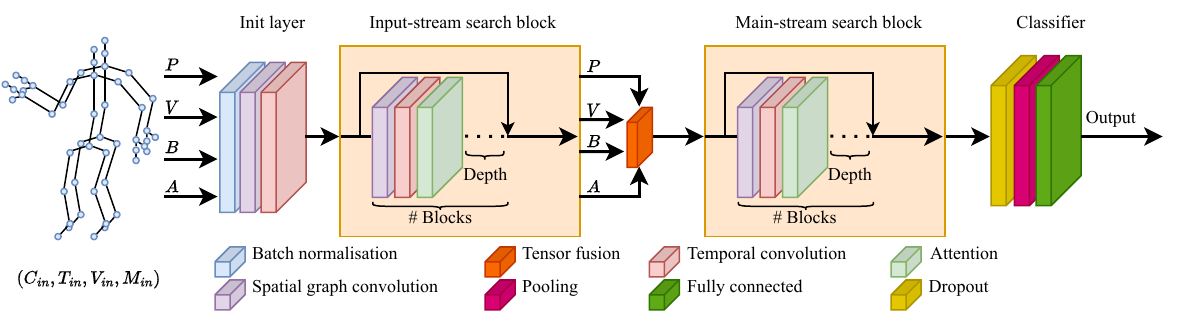}
    \caption{Architecture skeleton of the AutoGCN algorithm.}
	\label{fig:autogcn_arch}
\end{figure*}

\noindent As previously discussed, the architecture construction necessitates careful consideration of the unique data representation intrinsic to the skeleton sequence. To address this requirement effectively, the architecture is thoughtfully partitioned into separate building blocks, building upon prior research \cite{Song2022ConstructingSA, zhangSemanticsGuidedNeuralNetworks2020, xiaMultiScaleMixedDense2021}.
The fundamental elements comprise \textbf{1)} the initialization layer, \textbf{2)} the input stream- and, \textbf{3)} the main stream search block, followed by \textbf{4)} the classifier block, as depicted in Fig. \eqref{fig:autogcn_arch}.
Each block integrates various parameters derived from the comprehensive search space in Table \eqref{tab:search_space}. 
This modular approach ensures that the architecture can deal with the intricate nuances and complexities inherent in the skeleton sequence data representation.

\subsection{Search Algorithm}
AutoGCN seeks optimization on two fronts: the hyperparameters denoted as $h$, and the architecture represented as $\alpha$. 
This approach aligns with prior research in this field \cite{dongAutoHASEfficientHyperparameter2021, whiteNeuralArchitectureSearch2023}. 
The overarching objective is to minimize the loss function $\mathcal{L}$ on the validation dataset $\mathcal{D}_{val}$ while ensuring that the optimal parameters $\omega_{\alpha, h}^*$ are derived from the minimized loss on both the architecture $\alpha$, hyperparameters $h$ and the training dataset $\mathcal{D}_{train}$. 
This is articulated as follows:
\begin{equation}
    \min_{\alpha, h} \mathcal{L}(\alpha, \omega_{\alpha, h}^*, \mathcal{D}_{val})
    \quad \text{s.t.} \quad \omega_{\alpha, h}^* = \mathcal{L}(\alpha, h, \mathcal{D}_{train}).
\end{equation}

To address the potential inefficiencies of the Reinforce algorithm's sampling behavior, AutoGCN incorporates a replay memory $\mathbb{N}$ that stores a collection of $\mathcal{N}$ previously generated student architectures. 
This memory is used to enhance the sampling procedure for updating the controller. 
The addition of a replay memory helps mitigate the limitations of the Reinforce algorithm by utilizing a broader range of student architectures that have been saved over time. 
This has the effect of expanding the effective sampling size and improving the controller's decision-making process. 
Refer to Algorithm \eqref{alg:auto_gcn} for further details.

\begin{algorithm}[]
\label{alg:auto_gcn}
\DontPrintSemicolon
  
\KwIn{Split the data into: $\mathcal{D}_{\text{train}}$ and $\mathcal{D}_{\text{val}}$}
\KwOut{$\mathcal{P}_{max}(\alpha, h)$}

Initialize controller's policies $\mathcal{P}$, iterations $i$, and rollouts $r$\;

\While{not converged}{
	\For{\textup{i}}{
        
    		Sample $(\alpha, h)$ from controller's search space\;
    		Build student architecture\;
    		\For{\textup{25 epochs}}{
        		Train student architecture\;
       		Append validation accuracy to $r$
    		}
    }		
    Sample $\mathcal{N}$ student architectures from reservoir $\mathbb{N}$\;
    Update controller by REINFORCE with $r$\;
    Sample $\mathcal{P}_{max}(\alpha, h)$ and train architecture \;
}
Get final architecture from $\mathcal{P}_{max}(\alpha, h)$\;
\caption{AutoGCN algorithm} 
\end{algorithm}

\section{Experiments}
\noindent This section outlines our experimental setup of the proposed AutoGCN algorithm on two datasets: NTU RGB+D 60 and NTU RGB+D 120 \cite{shahroudyNTURGBLarge2016, liuNTURGB1202020}.
We compare current state-of-the-art (SOTA) models and the two introduced NAS procedures on these datasets. 
Furthermore, we investigate the impact of controller hyperparameters on the model's performance, including the number of rollouts and update cycles. 
Additionally, we perform experiments to assess the importance of the acceleration feature in our data representation and the influence of the searched network size.

\subsection{Datasets}
\noindent NTU RGB+D 60 is a substantial 3D human activity dataset for action recognition comprising two versions: NTU RGB+D 60 \cite{shahroudyNTURGBLarge2016} and NTU RGB+D 120 \cite{liuNTURGB1202020}. 
NTU RGB+D 60 contains 56,880 videos, each representing one of the 60 action classes.  
NTU RGB+D 120 is an extension of NTU RGB+D 60, including 60 additional action classes and 114,480 videos. 
To evaluate the classification performance, we follow the cross-subject and cross-view settings for NTU RGB+D 60 \cite{shahroudyNTURGBLarge2016}, and cross-subject and cross-setup settings for NTU RGB+D 120, as suggested respectively in \cite{liuNTURGB1202020, shahroudyNTURGBLarge2016}.

The NTU RGB+D 60 dataset is partitioned and used for evaluation as follows: 
(1) Cross-Subject (X-Sub) evaluation divides the dataset based on subjects. 
Specifically, 20 subjects are assigned to the training set, while the remaining 20 form the test set. 
(2) Cross-View (X-View) evaluation involves partitioning the dataset based on camera views. 
For this evaluation, camera views two and three are utilized to create the training data, while camera view one is reserved for testing.

Correspondingly, the following split protocols are suggested for NTU RGB+D 120 by the authors:
(1) Cross-Subject (X-Sub120): The training set comprises samples from 56 subjects, while the test set includes samples from 50 subjects.
(2) Cross-Setup (X-Setup120): In this protocol, samples with even setup IDs are designated for training, while samples with odd setup IDs are reserved for testing.

\subsection{Implementation}
In the experimental setup, the student architecture is trained for a maximum of 25 epochs, while the argmax architecture is trained for 80 epochs to ensure complete convergence. 
The learning rate used in training is sampled from the search space and undergoes a warm-up for the first ten epochs with a decay factor of $0.5$. 
Subsequently, the learning rate is reduced by $0.25$ at epochs 30, 50, 60, 65, and 70.
An early stopping mechanism is employed to identify underperforming student architectures during the initial training phase, terminating the training process before the sixth epoch if the accuracy is below 50\%.

The controller is updated with the Adam optimizer and a learning rate of $0.001$.
Once the predetermined number of student architectures have been sampled and trained, the controller is updated with those achieved accuracies as rewards. 
Simultaneously, sampling from the replay memory continues, and only student architectures achieving an accuracy higher than 80\% during the training process are added to the replay memory.

The training is performed on a single NVIDIA-V100 with 32 GB GPU RAM on the PyTorch framework (version 2.0.1) \cite{paszkePyTorchImperativeStyle2019} with the global seed set to $1234$. 
The code and the experiment results are publicly available at https://github.com/DeepInMotion/AutoGCN.

\subsection{Comparison with other SOTA}
To evaluate the performance of AutoGCN, it is compared with the baseline approach from Peng et al. \cite{dongAutoHASEfficientHyperparameter2021} and other state-of-the-art (SOTA) HAR approaches. 
The best-performing models' results on the NTU RGB+D 60 database are shown in Table \eqref{tab:sota_ntu60} and \eqref{tab:sota_ntu120}, respectively.
The values of $\mathcal{P}_{max}(\alpha, h)$ for the best-performing model are listed in Table \eqref{tab:found_sp}.
Moreover, the values of the Policies $\mathcal{P}$ for all search space parameters are listed in Appendix \eqref{appendix:x_sub} Fig. \eqref{fig:policy_xsub60} and Appendix \eqref{appendix:x_view} Fig. \eqref{fig:policy_xview60}, accompanied by an analysis of these values.

In contrast to other SOTA methodologies, we present the achieved point estimate for the NTU RGB+D 60 dataset and the associated confidence interval, calculated using bootstrap resampling \cite{efronIntroductionBootstrap1994}.
Specifically, we perform 1000 resamples of the complete test dataset with replacement while maintaining a constant training set and model configuration.
Our reporting is based on the boundaries encompassing the $[2.5, 97.5]$ percentiles, establishing a $95\%$ confidence interval for the test set.

As shown in Table \eqref{tab:sota_ntu60}, our approach surpasses the NAS-GCN baseline with the Joint configuration by 0.8\% for the X-View and the X-Sub dataset.
In contrast to SNAS-GCN, our model demonstrates a notable increase in accuracy, with a substantial gain of 1.2\% observed for both the X-Sub and X-View datasets.
Moreover, it is essential to emphasize that AutoGCN achieves similar results to the baseline method NAS-GCN without the need for ensemble techniques, achieving a competitive accuracy of 95.5\% on the X-View dataset. 
These results show the effectiveness and efficiency of AutoGCN in the context of skeleton-based human action recognition by delivering high performance without the added complexity of ensemble methods.

Table \eqref{tab:sota_ntu120} demonstrates that our approach yields results comparable to the SOTA models for the NTU RGB+D 120 dataset previously reported.

As delineated in Table \eqref{tab:found_sp}, a discernible disparity in the identified architectural components between the X-Sub and X-View datasets emerges. 
This disparity supports our contention from the introduction that these two datasets have inherent characteristics that cannot be effectively captured by a single searched architectural framework alone.

\begin{table}[]
    \caption{Comparison with SOTA models and NAS approaches on the NTU RGB+D 60 dataset with the Top-1 accuracy in (\%). The square brackets indicate the 95\% confidence intervals, determined via bootstrapping.}
    \label{tab:sota_ntu60}
    \centering
    \begin{tabular}{@{}lll@{}}
    \toprule
    Model           & X-Sub  & X-View \\ 
    \midrule
    ST-GCN \cite{yanSpatialTemporalGraph2018}                           &   81.5        &  88.3      \\
    AS-GCN \cite{liActionalStructuralGraphConvolutional2019}            &   86.8        &  94.2         \\
    GR-GCN \cite{gaoOptimizedSkeletonbasedAction2019}                   &   87.5        &  94.3 \\
    2S-AGCN \cite{shiTwoStreamAdaptiveGraph2019}                        &   88.5\footnotemark[1]        &  95.1\footnotemark[1]         \\ 
    AGC-LSTM \cite{siAttentionEnhancedGraph2019}                        &   89.2        &  95.0     \\  
    SGN \cite{zhangSemanticsGuidedNeuralNetworks2020}                   &   89.0        & 94.5 \\ 

    MST-AGCN \cite{huSkeletonMotionRecognition2022}                     & 89.5\footnotemark[1]      &   95.5\footnotemark[1]  \\
    EfficientGCN-B0 \cite{Song2022ConstructingSA}                       & 90.2                      &   94.9     \\ 
    STGAT (Joint) \cite{huSpatialTemporalGraph2022}                     & 90.2                      &   95.4        \\
    \midrule
    SNAS-GCN \cite{jiangSkeletonBasedHumanAction2023}                   & 89.0\footnotemark[1]                      & 95.0\footnotemark[1]  \\
    SNAS-GCN (Joint) \cite{jiangSkeletonBasedHumanAction2023}           & 87.1                                      & 94.3                         \\
    NAS-GCN \cite{pengLearningGraphConvolutional2020}                   & 89.4\footnote[1]{ensemble score}          & 95.7\footnotemark[1]            \\
    NAS-GCN (Joint) \cite{pengLearningGraphConvolutional2020}           & 87.5                                      & 94.7                         \\
    \midrule
    AutoGCN                                                             & 88.3 [86.9, 89.2]                         & 95.5  [94.6, 96.3]      \\ \bottomrule
    \end{tabular}%
\end{table}

\begin{table}[]
    \caption{Comparison with SOTA models on the NTU RGB+D 120 dataset with the Top-1 accuracy in (\%).}
    \label{tab:sota_ntu120}
    \centering
    \begin{tabular}{@{}lll@{}}
    \toprule
    Model                                                               & X-Sub120      & X-Setup120 \\ 
    \midrule
    ST-GCN \cite{yanSpatialTemporalGraph2018}                           &   70.7                    &  73.2      \\
    AS-GCN \cite{liActionalStructuralGraphConvolutional2019}            &   77.9                    &  78.5         \\
    2S-AGCN \cite{shiTwoStreamAdaptiveGraph2019}                        &   82.5\footnotemark[1]    &  84.2\footnotemark[1]         \\ 
    SGN \cite{zhangSemanticsGuidedNeuralNetworks2020}                   &   79.2                    & 81.5 \\ 
    EfficientGCN-B0 \cite{Song2022ConstructingSA}                       &  86.6                     &   85.0     \\    
    
    \midrule
    AutoGCN                                                             & 83.3      & 84.1       \\ \bottomrule
    \end{tabular}%
\end{table}

\begin{table}[htbp]
\caption{Comparison of the found architecture and hyperparameter components from the best performing models.}
\label{tab:found_sp}
\centering
    \begin{tabular}{@{}lll@{}}
        \toprule
        \textbf{Parameter}        & X-Sub & X-View \\ \midrule
        \multicolumn{2}{l}{\textbf{Common}}                                             \\ \midrule
        Activation layer        & Hardswish & Hardswish     \\
        Attention layer         & Fa        & Ja            \\
        Conv. layer type        & Bottleneck        & SG    \\
        Dropout probability     & 0.25      & 0.15          \\
        Init layer size         & 64       & 256            \\ \midrule
        \multicolumn{2}{l}{\textbf{Input stream}}                                             \\ \midrule
        Blocks in               & 1         & 2     \\
        Depth in                & 2         & 1     \\
        Stride in               & 3         & 3     \\
        Scaling factor	        & 0.4       & 0.6   \\
        Temporal window         & 5         & 3     \\
        Graph distance          & 3         & 2     \\
        Reduction factor        & 1.25      & 1.35  \\ \midrule
        \multicolumn{2}{l}{\textbf{Main stream}}                                              \\ \midrule
        Blocks main             & 3         & 3     \\
        Depth main              & 1         & 3     \\
        Graph distance          & 11         & 9    \\
        Shrinkage               & 2         & 2     \\
        Residual layer          & True      & True  \\
        Adaptive                & True      & True  \\ \midrule
        \multicolumn{2}{l}{\textbf{Optimizer}}                                                \\ \midrule
        Optimizer               & SGD     & AdamW       \\ 
        Learning rate           & 0.05      & 0.01      \\
        Weight decay            & 0.001       & 0.0     \\
        Momentum     	        & 0.5       & 0.5        \\
        Batch size    	        & 16         & 8         \\ \bottomrule
    \end{tabular}
\end{table}

\subsection{Comparison to Random Search}
Random search \cite{bergstraRandomSearchHyperparameter2012} as an optimization technique for tuning hyperparameters is widely used in NAS procedures, which makes it an essential component of our experiments \cite{lindauerBestPracticesScientific2020}. 
To ensure the fairest comparison against our proposed method, we perform the random search following the same steps as our algorithm AutoGCN: An initial cohort of 20 student architectures are trained for 25 epochs, and subsequently, the highest-performing architecture is chosen and trained for an extended training period of 80 epochs.
Finally, we compare the iterations taken by each approach to obtain the best-performing architectures, displayed in Table \eqref{tab:random_search}.
It can be observed that the AutoGCN algorithm achieves a higher-performing architecture after just 20 iterations both for the X-View and X-Sub datasets compared to random search.
To achieve a comparable point-estimates as AutoGCN, random search has to undergo another 20 iterations.
The results indicate that AutoGCN can obtain an optimal architectural configuration without being subject to the stochasticity of random search.

\footnotetext[1]{ensemble score}

\begin{table}[]
\centering
\caption{Comparison of random search and AutoGCN to the period of time and the Top-1 accuracy in (\%).}
\label{tab:random_search}

\begin{tabular}{@{}lll@{}}
\toprule
 & X-Sub & X-View \\ \midrule
Random search           &       86.8 - 20 iterations   &  92.9 - 20 iterations                                 \\
AutoGCN                 &       \textbf{88.3 - 20} iterations   &  \textbf{95.1 - 20} iterations               \\  \bottomrule
\end{tabular}%

\end{table}


\subsection{Influence of the Controller hyperparmeters}
\begin{figure*}[!hbt]
	\centering
	\includegraphics[width=\textwidth]{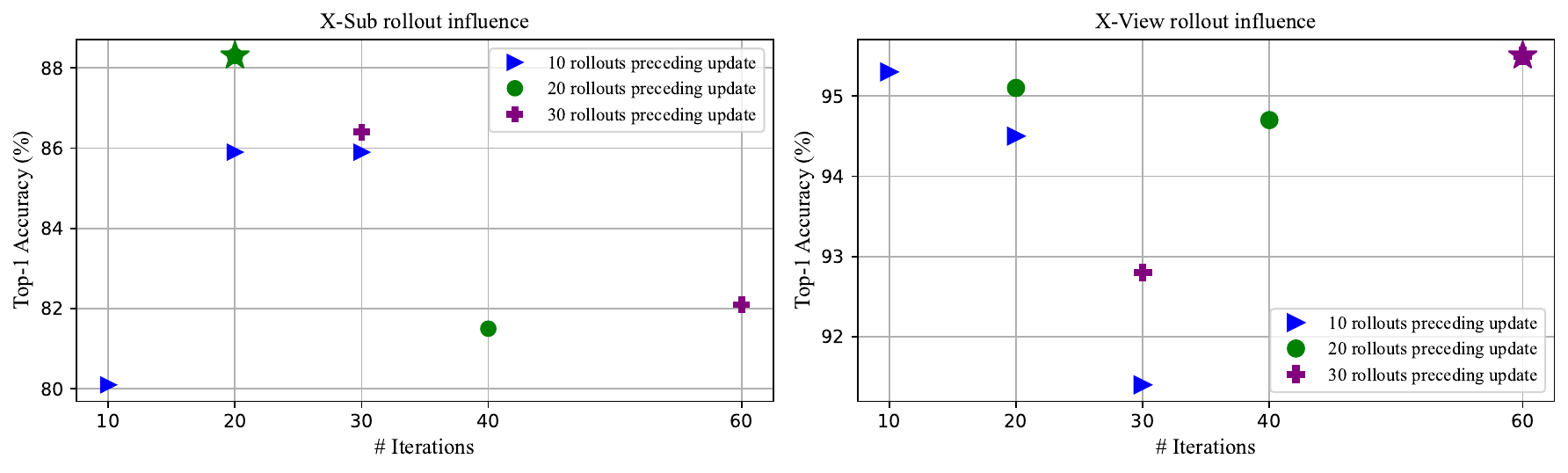}
    \caption{Influence of the rollout parameter on the model performance. The star indicates the highest accuracy achieved in the rollout experiments.}
	\label{fig:autogcn_resul}
\end{figure*}

Since the controller has fixed hyperparameters, we investigate the ideal number of rollouts for training our model, aiming to balance computational efficiency and the final model performance. 
The found values are shown in Table \eqref{tab:contr_hyper}, in which the Top-1 accuracy percentages in relation to the rollout hyperparameter of the controller and the update stages are depicted. 
It also displays the average accuracies of the student models and Top-1 accuracies for the first, second, and third updates in percentage values. 
The table provides an overview of how the choice of rollouts impacts the accuracy of the two datasets, X-Sub and X-View. 
The arrow symbols describe the trend in the accuracy and average accuracy compared to the previous update: '↑' for an increase, '↓' for a decrease, '→' for no significant change.
Furthermore, those results are visualized in Fig. \eqref{fig:autogcn_resul}.

\begin{table}[]
\centering
\caption{Top-1 accuracy percentages w.r.t the rollout hyperparameter of the controller and the update states, along with the average accuracies of the student models and Argmax accuracies for the update cycles in (\%). The table showcases how the choice of rollouts impacts the accuracy of two distinct approaches, X-Sub and X-View, and the effects of the first, second, and third updates on their respective performance.}
\label{tab:contr_hyper}
\begin{tabular}{llllll}
    \toprule
    \textbf{Rollouts}       & Top-1         & $1^{st}$ update   & $2^{nd}$ update   & $3^{rd}$ update      \\
                            &               &   Avg. - Top-1    & Avg. - Top-1      & Avg. - Top-1                \\
    \midrule
    \multicolumn{4}{l}{\textbf{X-Sub}}                                             \\
        \midrule
        10 & 85.9              & 50.4 - 80.1            & 61.3 $\uparrow$ - \textbf{85.9} $\uparrow$        & 62.3 $\uparrow$ - 85.9 $\rightarrow$  \\
        20 & \textbf{88.3}     & 45.7 - \textbf{88.3}   & 33.5 $\downarrow$ - 81.5  $\downarrow$            & --                                     \\
        30 & 86.4              & 49.6 - \textbf{86.4}   & 60.3 $\uparrow$ - 82.1 $\downarrow$               & --                                    \\
        \midrule
    \multicolumn{4}{l}{\textbf{X-View}}                                             \\ 
    \midrule
        10 & 95.3               & 71.3 - \textbf{95.3}      & 72.0 $\rightarrow$ - 94.5 $\downarrow$            & 68.6 $\downarrow$ - 91.4 $\downarrow$        \\
        20 & 95.1               & 56.2 - \textbf{95.1}      & 66.4 $\downarrow$ - 94.7 $\rightarrow$             & --         \\
        30 & \textbf{95.5}      & 69.1 - 92.8               & 60.0 $\downarrow$ - \textbf{95.5} $\uparrow$      & --         \\
    \bottomrule
\end{tabular}
\end{table}

The X-View dataset attains its highest point estimate following 30 rollouts after the second update cycle. 
Notably, for 30 rollouts, the controller identifies the best-performing architecture after the second update, requiring significantly more time than the experiment with only ten or 20 samples between updates while not achieving a notable higher point estimate.

Conversely, the X-Sub dataset achieves its highest point estimate after 20 rollouts, which also transpires following the first update cycle. 
The X-Sub dataset necessitates only one update cycle to achieve the highest point estimate for 20 and 30 rollouts.
With ten rollouts, the controller achieves the highest accuracy after the second iteration and remains constant afterward.
With 30 rollouts, the controller reaches a state where it becomes more confident about a suboptimal model configuration, resulting in diminished performance.

Given that the student architectures are trained for a shorter duration of 25 epochs, the optimizer's hyperparameters significantly impact the final accuracy used as a reward for the controller.
With a ``fast learning'' optimizer, suboptimal student architectures, which converge more rapidly, may enjoy an advantage over ``slower'' trained architectures in the early stages of training since those are only trained for 25 epochs.

Considering that the selection of the student architecture is stochastic, the controller may be updated with less performant models between each update cycle. 
This becomes evident when comparing the average accuracy for the X-Sub dataset with ten rollouts, where a notable rise in the average accuracy correlates with an increase in the highest point estimate.
On the other hand, the second update with 30 rollouts decreases the point estimate for the X-Sub datasets, suggesting that the controller becomes trapped in a suboptimal configuration after this amount of updates.
Furthermore, it has to be recognized that the search space contains multiple optima that can be explored, resulting in the alteration of optimal architecture builds.
Consequently, different architecture configurations can achieve similar performances after optimization, which becomes evident when investigating the different sampled architecture configurations across the sampled rollouts.

\subsection{Influence of the model size}
We conduct experiments to investigate the optimal network size by altering the search space, as outlined in Table \eqref{tab:search_space}. 
These focus on exploring larger architectures, where we increased layers, blocks, and depth sizes in the experimental design.
\begin{table}[]
\caption{Changed values of the search space and the resulting point estimate in (\%) on the X-View dataset.}
\label{tab:big_sp}
\centering
    \begin{tabular}{@{}llll@{}}
        \toprule
        \multicolumn{3}{l}{\textbf{Search Space}}       \\ \midrule
        Init layer size         &   \multicolumn{2}{l}{[156, 195, 256, 320]}  \\ 
        Blocks in               &   \multicolumn{2}{l}{[2, 3, 4]}            \\
        Depth in                &   \multicolumn{2}{l}{[3, 4, 5]}            \\                                  
        Blocks main             &   \multicolumn{2}{l}{[5, 6, 7, 8]}         \\  
        Depth main              &   \multicolumn{2}{l}{[3, 4, 5]}            \\ \midrule
        \textbf{Performance}    & FLOPs         &  Parameters   & Top-1            \\ \midrule
        Large model             & 13.94G        &   4.79M       & 95.4              \\
        Small model             & 14.47G        &   1.76M       & 95.3              \\
        \bottomrule
    \end{tabular}
\end{table}
The results in Table \eqref{tab:big_sp} indicate that the X-View dataset's accuracy remains consistent for the larger and smaller sampled architecture after 20 rollouts.
The small model requires much fewer parameters but slightly more FLOPs than the larger model.
The greater amount of FLOPs is due to the more complex \textit{Bottleneck} convolutional layer chosen by the controller.

This outcome shows the effectiveness of the proposed search space, as defined in Table \eqref{tab:search_space}, and emphasizes the usage of smaller models that require significantly less training. 
The results advocate utilizing these smaller models for their computational efficiency while achieving the same performance as the large model, requiring fewer computational resources.

\subsection{Data representation}
The newly introduced acceleration feature is analyzed to gauge the strength of the data representation.
Consequently, this feature is excluded and incorporated into the highest-performing model identified through our search. 
The impact of omitting this feature from the model becomes evident when observing the results presented in Table \eqref{tab:accel_abl}, where the Top-1 accuracy demonstrates a notable decrease.
When conducting a two-sided z-test with a 5\% significance level, the null hypothesis of no significant difference between those data modalities can be rejected.

\begin{table}[]
\centering
\caption{Influence of the acceleration feature $A$ on the Top-1 accuracy in (\%) with the 95\% confidence interval.}
\label{tab:accel_abl}
\begin{tabular}{llllll}
    \toprule
    Input & \textbf{X-Sub} & \textbf{X-View} \\
    \midrule
    $P, V, B$        & $87.5 \pm 1.4$       & $93.7 \pm 1.1$    \\            
    $P, V, B, A$     & $88.3 \pm 0.9$       & $95.6 \pm 0.8$    \\  
    \bottomrule
\end{tabular}
\end{table}

\section{Limitations \& Future Work}
The proposed AutoGCN algorithm uses a reinforcement learning approach, requiring every student network to be trained from scratch at each iteration.
While this strategy enables dynamic adaptation and refinement of the individual architecture and hyperparameter components, it also leads to an increased computational burden in computing the reward for the controller guiding the search and updating the policies $\mathcal{P}$.

In order to avoid having to train every student architecture from scratch, the search procedure could adopt so-called \textit{one-shot} techniques like DARTs, in which an over-parameterized \textit{supernetwork} is trained to contain every potential architecture build and enable weight sharing among students \cite{whiteNeuralArchitectureSearch2023, liuDARTSDifferentiableArchitecture2019}.
To implement this approach, reusable building blocks, and operation types that can be recycled among different student compositions effectively would have to be defined.
While such an adaptation could then lead to a speed-up of the search procedure, it could also lower the versatility of the search space.

In our work, speed-up techniques for the search process are limited to leveraging the knowledge reservoir $\mathcal{N}$ and implementing an early-stopping mechanism.
Possible enhancements to accelerate the search could involve the inclusion of performance predictors \cite{whiteNeuralArchitectureSearch2023}.
Such predictors could assess the potential of a given student architecture early, allowing for a more efficient allocation of computational resources by prioritizing the exploration of promising candidates and neglecting unpromising ones.
With such techniques, the convergence towards optimal solutions could be accelerated, resulting in a significantly reduced overall search time.


\section{Conclusion}
In this study, we have developed and presented a GCN NAS algorithm named AutoGCN, tailored for the task of skeleton-based HAR. 
The intricate dependencies between hyperparameters and architectural configurations are formulated in an expressive search space encompassing a broad range of building blocks from which the controller can sample, which allows the algorithm to find a versatile and high-performing network architecture.
AutoGCN is applied to identify and optimize these fundamental building blocks of the network concurrently with a reinforcement algorithm, giving every search space parameter a policy that can be updated based on the anticipated performance from the sampled student architectures.
The search process's exploration and exploitation behavior is refined by incorporating a replay memory during the search process, enabling the method to strike a promising balance between these and enhancing the overall search performance.

Through extensive experiments, we provide a rigorous performance analysis, comparing our method against the baseline NAS procedure and SOTA approaches in the domain of skeleton-based HAR.
Finally, AutoGCN demonstrates effectiveness also compared to random search.

Future work will investigate the influence of performance predictors on AutoGCN to decrease the search time of the algorithm, as good-performing architectures could be recognized earlier.
Additionally, weight-sharing techniques could further make the search more efficient.

\printbibliography

\appendix
\section{Policy values on the X-Sub dataset}\label{appendix:x_sub}
\begin{figure*}[htb]
    \centering
    \includegraphics[width=\textwidth]{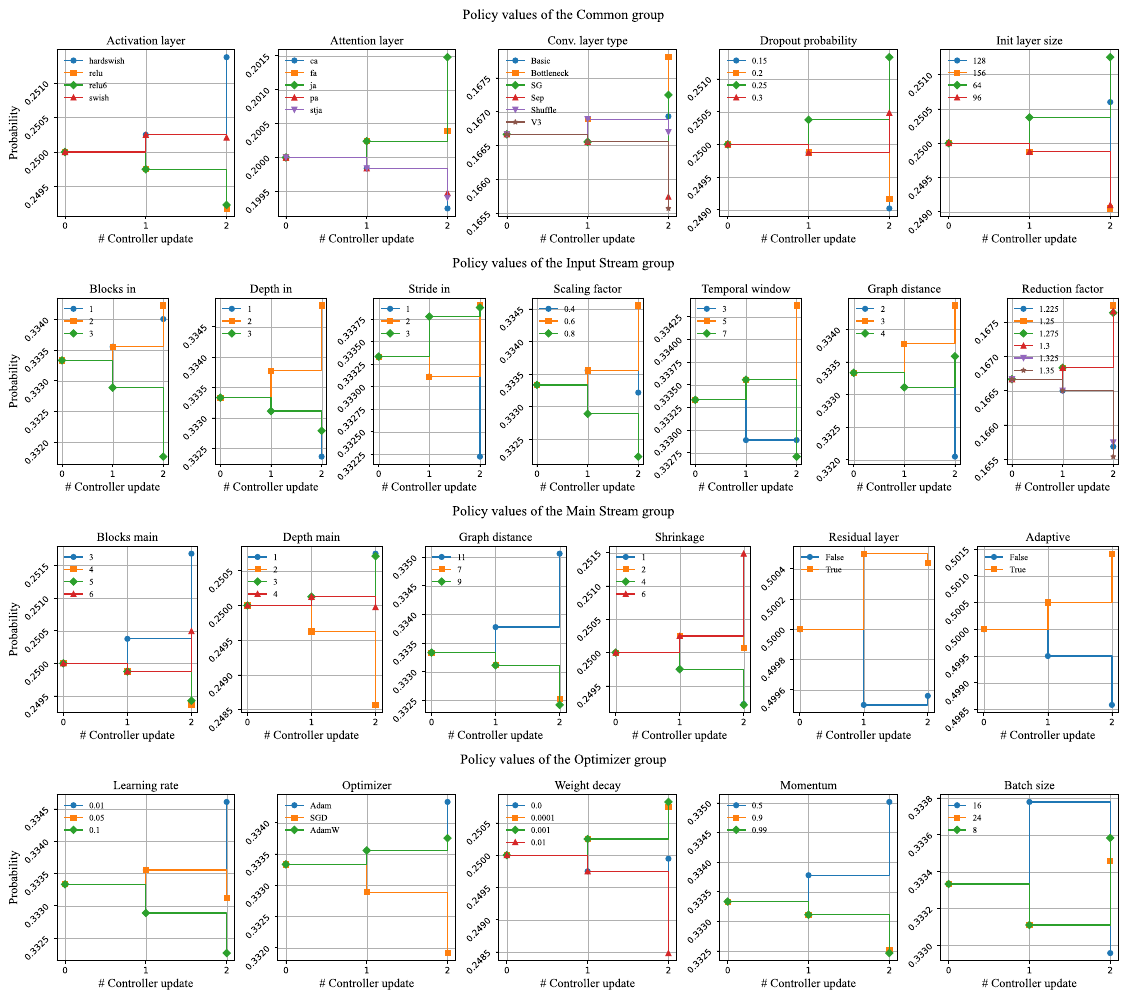}
    \caption{Policy values for the best-performing model on the X-Sub dataset. One update cycle contains 20 student architectures.}
    \label{fig:policy_xsub60}
\end{figure*}

Fig. \eqref{fig:policy_xsub60} shows the policy values for the model's hyperparameters and architecture search space with 20 rollouts between controller updates on the X-Sub dataset. 
These values are categorized within the four search areas as shown in Table \eqref{tab:search_space}. 
The highest Top-1 accuracy in this experiment is attained with the first controller update, with the final $\mathcal{P}_{max}(\alpha, h)$ values outlined in Table \eqref{tab:found_sp}.

Within the \textit{Common} group, most policy values maintain consistency after the second update cycle. 
However, within the \textit{Conv. layer} choices, uncertainty arises, particularly evident in the fluctuation of probabilities between the \textit{Basic} and \textit{Shuffle} layers after the first controller update. 
A similar trend of uncertainty is visible in the \textit{Dropout probability} and \textit{Init layer size}.
Regarding the \textit{Input stream} group, discernible trends emerge after both the first and second update cycles. 
The search space parameters with lower probability values tend to retain their trend after the first update.
Nonetheless, the controller is uncertain regarding \textit{Blocks in}, \textit{Stride in}, and the \textit{Reduction factor}, with a tight distribution across these search space parameters for the highest values.
In the \textit{Main stream} group, most search areas demonstrate clear trends, barring the values associated with \textit{Depth main}. 
These probability values for the depths of one or three layers remain closely clustered following the first and second controller updates.
Within the \textit{Optimizer} group, variability among the policy values is notable, particularly for the \textit{Learning rate} and \textit{Batch size} parameters. 
Given the intertwined nature of these hyperparameters and their impact on the incompletely trained student architecture, fluctuations are expected.

It is essential to highlight that multiple local optima are possible within this search space configuration. 
These optima depend upon the statistical variance from the random sampling process of the student architectures.
Furthermore, the average student's Top-1 accuracy is significantly lower with the second update cycle in this experiment compared to the other rollout experiments from Table \eqref{tab:contr_hyper}.

\section{Policy values on the X-View dataset}\label{appendix:x_view}
\begin{figure*}[htb]
    \centering
    \includegraphics[width=\textwidth]{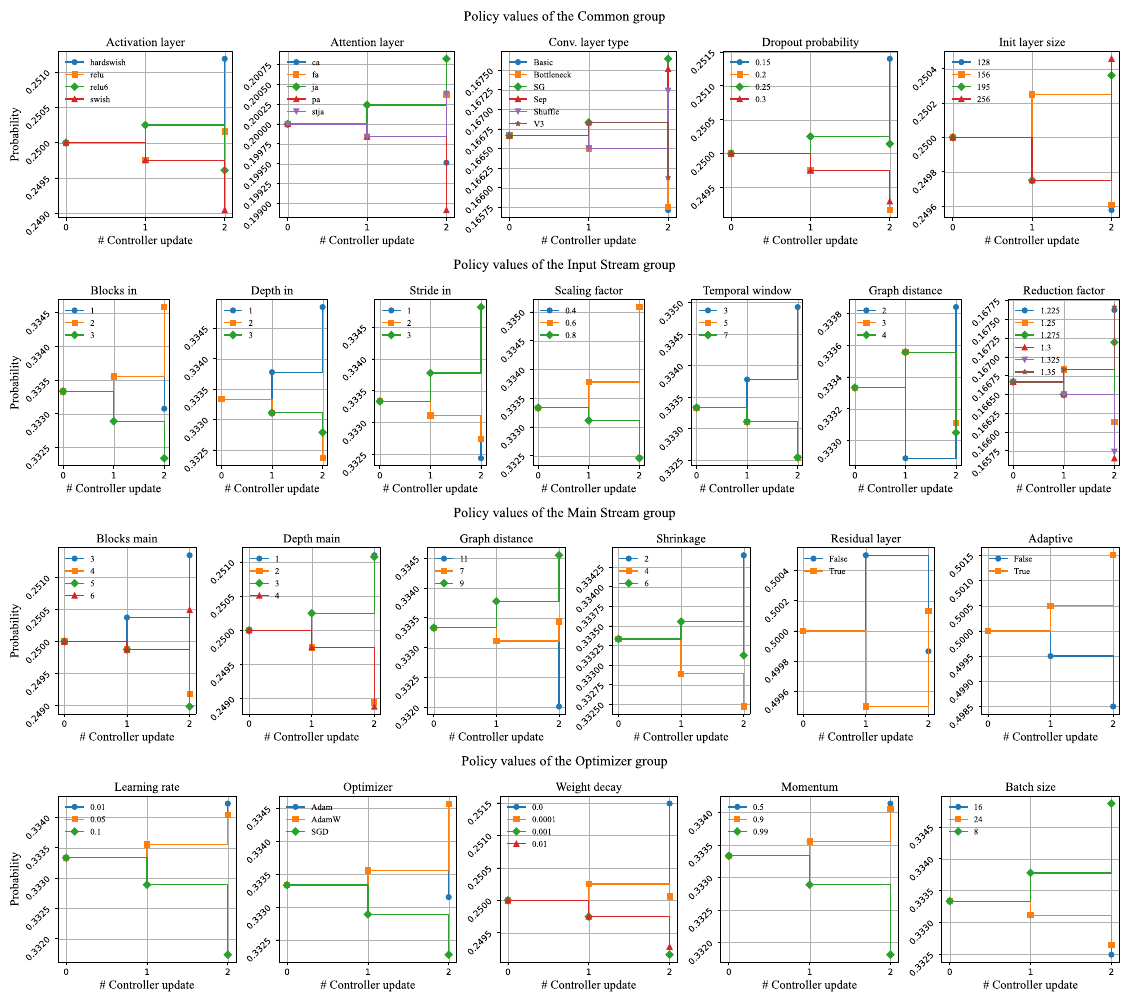}
    \caption{Policy values for the best-performing model on the X-View dataset. One update cycle contains 30 student architectures.}
    \label{fig:policy_xview60}
\end{figure*}

Fig. \eqref{fig:policy_xview60} shows the policy values for the model's hyperparameter and architecture search space with 30 rollouts between controller updates on the X-View dataset. 
The values are grouped as defined in Table \eqref{tab:search_space}. 
The best Top-1 accuracy in this experiment is achieved with the second controller update, where the $\mathcal{P}_{max}(\alpha, h)$ values are presented in Table \eqref{tab:found_sp}.

For the \textit{Common} group, the trend from the first to the second controller update is relatively consistent in its upward direction for the highest search space value.
In contrast, the final value for the \textit{Init layer size} is experiencing a significant increase in its probability value.
The policy values with lower probabilities fluctuate more between the first and second controller updates.
In the \textit{Input stream} group, most policy values indicate a clear trend for the one single search space value.
In contrast, the \textit{Graph distance} and the \textit{Reduction factor} show variability between the first and the second update.
Regarding the \textit{Main stream} group, most search areas manifest clear trends, except for the values associated with \textit{Depth main}. 
As in the X-Sub experiment, the probability values for depths of one or three layers remain closely clustered following the first and second controller updates.
Additionally, the values for the \textit{Residual layer} converse after the second update.
In the \textit{Optimizer} group, the policy values indicate less fluctuation than for the X-Sub dataset in Fig. \eqref{fig:policy_xsub60}.
A clear trend among these different groups is visible from the first to the second controller update.

As with the X-Sub experiments, the search space values contain a statistical variance due to the random sampling process of the student architectures.

\end{document}